%% file: sample_FG2023.tex
\documentclass[a4paper, 10pt, conference]{ieeeconf}      
\usepackage{FG2023}

\IEEEoverridecommandlockouts                              
\usepackage{amsmath}
\usepackage{graphicx}
\usepackage{float}
\usepackage{authblk}
\usepackage{graphicx}
\usepackage{multicol}
\usepackage{multirow}
\usepackage{colortbl}
\makeatletter
\let\NAT@parse\undefined
\makeatother
\usepackage[colorlinks=true, allcolors=blue]{hyperref}

\title{Analysis of Human Perception in Distinguishing \\ Real and AI-Generated Faces: An Eye-Tracking Based Study}

\author[1]{Jin Huang~\thanks{This work was done during an internship at Dolby Laboratories.}}
\author[2]{Subhadra Gopalakrishnan}
\author[2]{\\Trisha Mittal}
\author[2]{Jake Zuena}
\author[2]{Jaclyn Pytlarz}

\affil[1]{University of Notre Dame, Department of Computer Science and Engineering}
\affil[2]{Dolby Laboratories, Advance Technology Group}

\begin{document}

\ifFGfinal
\thispagestyle{empty}
\pagestyle{empty}
\else
\pagestyle{plain}
\fi
\maketitle

\begin{abstract}

Recent advancements in Artificial Intelligence have led to remarkable improvements in generating realistic human faces. While these advancements demonstrate significant progress in generative models, they also raise concerns about the potential misuse of these generated images. In this study, we investigate how humans perceive and distinguish between real and fake images. We designed a perceptual experiment using eye-tracking technology to analyze how individuals differentiate real faces from those generated by AI. Our analysis of StyleGAN-3 generated images reveals that participants can distinguish real from fake faces with an average accuracy of $76.80\%$. Additionally, we found that participants scrutinize images more closely when they suspect an image to be fake. We believe this study offers valuable insights into human perception of AI-generated media.



\end{abstract}

\input{sections/0_introduction}
\input{sections/1_related_work}
\input{sections/2_study}
\input{sections/3_analysis}

\input{sections/5_conclusion}

{\small
\bibliographystyle{ieee}
\bibliography{egbib}
}

\clearpage

\end{document}

%% file: sections/0_introduction.tex
\section{Introduction}\label{intro}

In recent years, advancements in face synthesis have been nothing short of remarkable, fueled by the convergence of deep learning techniques and vast datasets. One notable breakthrough comes from Generative Adversarial Networks (GANs), exemplified by StyleGAN and its subsequent iterations~\cite{karras2019style, karras2020analyzing, karras2022stylegan3}. These models have become a powerful tool for generating photorealistic human face images by learning the distribution of real face images in a dataset.

These advancements have offered great benefits with their many applications by enabling unprecedented levels of creativity and realism in various fields such as virtual reality~\cite{entertainment} and digital art~\cite{digital_art}. However, they have also raised ethical and privacy concerns. The ability to generate realistic fake faces could be exploited for malicious purposes, including identity theft, misinformation campaigns, and deepfake videos that threaten individuals' reputations and manipulate public opinion~\cite{AGARWAL2023223, identity_theft, fake_news}. Furthermore, there are concerns regarding consent and the unauthorized use of individuals' likeness, highlighting the need for robust regulations and safeguards to mitigate potential risks.

Given these benefits as well as concerns, understanding how humans perceive real and fake face images is crucial. 
By gaining insights into clues that differentiate real faces from synthetic ones, we can better equip ourselves to detect misinformation and fraudulent activities facilitated by increasingly sophisticated face synthesis tools. Besides, understanding human perception can help the development of ethical guidelines and regulatory frameworks to ensure the responsible use of face synthesis technology and safeguard individuals' privacy. Ultimately, by understanding how humans perceive real and fake face images, we can harness the full potential of face synthesis technology while mitigating its potential risks and pitfalls.

\begin{figure}[t]
\includegraphics[width=0.45\textwidth]{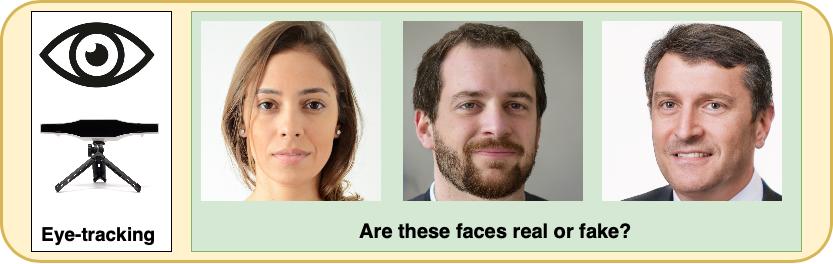}
\centering
\caption{\textbf{Analyzing Human Perception in Distinguishing Real and AI-Generated Faces:} We designed a perceptual experiment using eye-tracking to analyze human ability on distinguishing real and fake face images.}
\label{teaser}
\end{figure}
Motivated by these issues and the importance of understanding human perception, we studied how humans perceive real and fake face images in this work. Distinguishing real and fake face images is a very challenging task for humans for multiple reasons. First of all, rapid advancements in face synthesis algorithms have led to the creation of synthetic faces that closely resemble real ones, making it challenging to discern the differences. Besides, the immense variability in real faces, including differences in age, ethnicity, and facial expressions, can make it harder to establish a definitive standard for what constitutes a ``real" face. Moreover, the context in which the face images are presented can influence perception. For example, human perception can be impacted by the portrait image’s background, degree of image blur, and other non-facial features presented in the real and fake images~\cite{DBLP:journals/corr/abs-2111-04230}.

In this work, we studied how humans perceive real and fake images with eye-tracking. While there exists different methods to measure human perception, there are several advantages of using eye-tracking. First of all, eye-tracking provides precise measurements of location and duration when individuals look at specific areas of an image, allowing us to pinpoint which features are salient in distinguishing real from fake images. Based on this, it could be possible to gain insights into the attention patterns of individuals when viewing real and fake images, shedding light on which regions in an image attract the most attention. Most importantly, eye-tracking is a natural and intuitive method of gathering data on human visual behavior, requiring minimal effort from participants. Unlike other techniques that may involve complex training procedures or active participation, eye-tracking simply involves looking at stimuli displayed on a screen, which is a fundamental aspect of human behavior. These advantages allow us to collect rich data on gaze patterns without much burden.

To understand the human perception of synthetic faces, we utilized real face images sampled from Flickr-Faces-
HQ Dataset (FFHQ)~\cite{karras2019style} and fake face images generated by StyleGAN-3~\cite{karras2022stylegan3}, and studied to what extent humans can correctly distinguish real and fake face images. During the study, participants' responses were recorded, and at the same time, their gaze information was recorded by an eye-tracker. We conducted an in-depth analysis on the data we collected.

In summary, the contributions of our work are as follows:

\begin{itemize}
    
    \item We collected human responses and eye-tracking data on over $7,000$ images, which largely exceeded previous work where eye-tracking was engaged.

    \item We conducted an in-depth analysis on the data we collected, including but not limited to: recognition accuracy, reaction time, human gaze fixation, duration, gaze convex hull, gaze entropy and how they are related with human responses. Moreover, we explored how different non-facial features (background, accessory, etc) impacted human decisions.
\end{itemize}

For the rest of this paper, we first review some literature describing Generative Algorithms for face synthesis, and human behavioural study on AI generated contents in Section~\ref{related_work}. Then we describe our study protocol and experiment design in Section~\ref{design}. We give detailed analysis on the data we collected and present our results in Section~\ref{analysis}. Finally, we conclude our work by doing some discussion on our findings as well as providing some insights for future work in Section~\ref{conclude}.

%% file: sections/1_related_work.tex
\section{Related Work} \label{related_work}

\subsection{Generative Algorithms for Faces}

Recent image synthesis techniques encompass a variety of generative algorithms designed to maximize realism and diversity among the generated images. A subset of image generation works include work on controllable and realistic human face generation. For example, Generative Adversarial Networks (GANs)~\cite{goodfellow2014generative} are prominent for their ability to produce high-quality faces by learning distributions from large datasets~\cite{karras2019style}. Methods involving Variational Autoencoders (VAEs)~\cite{kingma2022autoencodingvariationalbayes} focus on latent space representations to generate novel faces. And most recently, many text-to-image models have been proposed and can be utilized for face synthesis, including but not limited to DALL-E~\cite{ramesh2021zeroshottexttoimagegeneration}, Stable Diffusion~\cite{rombach2022highresolutionimagesynthesislatent} etc. 


Built on the original GAN, conditional GAN~\cite{mirza2014conditional} incorporated additional conditioning information to control the generated output and enabled the modification of specific attributes of faces (e.g., age, gender, expression) while preserving other characteristics. Deep Convolutional GANs (DCGAN)~\cite{radford2015unsupervised} utilized deep convolutional neural networks (DCNNs) to the GANs in order to generate high-quality images. To tackle the difficulty for GANs to generate high resolution face images, Karras et al. proposed ProGAN~\cite{karras2017progressive}, which incrementally grew the resolution of generated images during training in the generator, using multi-resolution discriminators and smooth transitions between levels. Followed the idea of ProGAN, StyleGAN series~\cite{karras2019style, karras2020analyzing, karras2022stylegan3, antipov2017training} focused on generating high-quality, diverse, and controllable images with a strong emphasis on training stability and image fidelity, by adapting ``style" into training and eliminating aliasing artifacts.


\subsection{Human Behavioral Study on AI Generated Contents} 

As generative AI experiencing a surge in popularity in recent years, researchers have been shifting their attention to conducting human psychophysical studies and understanding human dynamics in the context of generated contents. Some work have been done to either use interactive software or physiological measurements to explore subjects' responses to real and fake contents. For example,~\cite{lago2021more} presented a study that measured the human ability to distinguish between real and fake face images from three GAN models by asking the participants to choose the authenticity of an image in a scale from $1$ to $7$.~\cite{Boyd_2023_WACV} collected human responses on whether they think a face image is real or fake, and obtained human saliency maps for fake images by asking the participants to label regions; the authors then incorporated these saliency maps into model training. ~\cite{liao2022perceptual} presented a dataset with a variety of translucent appearances; it also conducted human psychophysical experiments and asked observers to judge whether an object in the image was real or fake as well as rating the level of translucency of the material. These work, although successfully obtain human psychophysical responses reasonably, are rather mechanical than natural.

\subsection{Human Behavioral Study with Eye-tracking}

Another direction that researchers have been paying attention to is human eye-tracking, as it can provide real-time information on humans' natural behaviors and recognition dynamics. Several works have been done to conduct studies on human eye-tracking with respect to real and fake contents. For instance,~\cite{hansen2020factuality} studied the possibility of inferring whether a news headline was true or false using human eye movements and~\cite{bozkir2022regressive} studied human eye movements when reading fake news and real news and analyzed eye-tracking patterns.~\cite{simko2019fake} proposed a two-step approach where the authors asked participants to casually read news where some of the articles were fake, and in a second run, they asked the participants to decide on whether the news were real or fake; participants' eye-tracking information was recorded during the whole process.~\cite{brockinton2022utilising} conducted a study with online eye-tracking to assess the impacts of different cultural backgrounds on fake and real news decision-making. As for videos,~\cite{gupta2020eyes} presented an eye-tracking database to understand human visual perception of fake videos, and evaluated the ability of participants to detect fake video artifacts.

\begin{figure*}[t]
\includegraphics[width=0.8\textwidth]{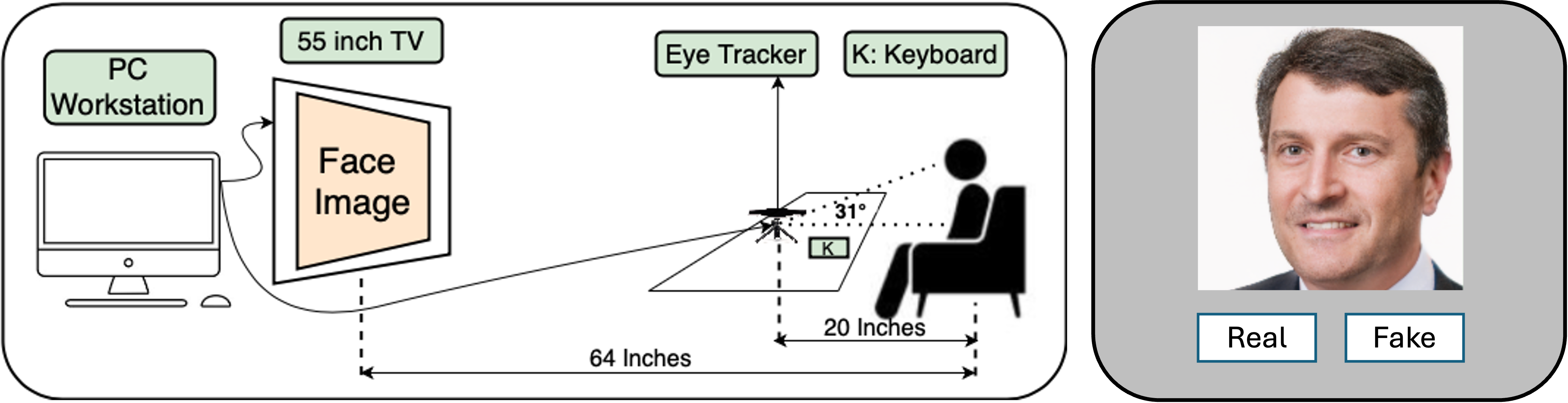}
\centering
\caption{\textbf{Experiment Setup: }\textit{(left)} Participants were seated 64" away from the TV screen (55 ") that was used to show them stimuli (real/fake face images). The eyetracker was placed 20" from the participant at an angle of 31 degrees. Participants used a keyboard for the experiment. \textit{(right)} A mock-up of the stimulus presentation to participants.}
\label{study_setup}
\end{figure*}

Our work went down the line of analyzing human eye-tracking data, but we focused on images instead of fake news or videos. Work that is most close to ours is~\cite{caporusso2020using}, where the authors investigated the perception of images produced by GANs and focused on individual’s ability to discriminate between fake and real profiles. They utilized eye-tracking to identify the presence of patterns in subjects’ gaze. Our work differs from this previous work in several ways. 

\begin{enumerate}
    \item We used StyleGAN-3~\cite{karras2022stylegan3} to generate our fake images, which is the state-of-the-art GAN model for human face generation, and collected human information on over $7,000$ images, which largely exceeded the $2134$ images in previous work.

    \item We conducted a much more thorough data analysis that was centered with human eye-tracking metrics, focusing on human gaze fixation, fixation duration, gaze convex hull, gaze entropy and how they were related with human decisions. 

    \item Our work showed that humans were pretty good at distinguishing real and fake face images generated by GAN, which differs from what was found in~\cite{karras2022stylegan3}.
\end{enumerate}  

%% file: sections/2_study.tex
\section{Study Description} \label{design}

In this section, we describe a few important components for our human perception study.

\subsection{Equipment}

In our study, we utilized a commercial eye-tracking device, named GazePoint GP3 HD ~\footnote{https://www.gazept.com/product/gp3hd/} (referred as ``the eye-tracker" in the rest of this article). 
It contains a machine-vision camera in its imaging and processing system, and captures human eye gaze at a frequency of 60 frames per second. 
We obtain gaze data from the eye tracker as a set of $X$ and $Y$ co-ordinates of the observers gaze location with respect to the screen. This gaze data is then processed by the Gazepoint software ~\footnote{https://www.gazept.com/blog/visual-tracking/eye-tracking-software-features-to-utilize/?v=7516fd43adaa} to obtain gaze fixations, and fixations are defined as the periods of time where the eyes are relatively still.~\footnote{https://connect.tobii.com/s/article/understanding-tobii-pro-lab-eye-tracking-metrics?language=en\_US}

The eye-tracker also comes with a software that can be used to help adjust its location and angle. This ensures that the participant’s eyes are in range with the camera. 
Besides the eye-tracker, we also used a desktop computer with Windows 10 operation system installed, a keyboard and mouse for this study.

 \subsection{Study Design and Setup}

At the beginning of the experiment, participants were invited to a lab where they were seated in a chair in front of a TV screen, and the eye tracker, keyboard and mouse were placed in between the participant and TV on a desk. Lights in the lab were adjusted to the minimal level to decrease the reflections on the TV. Participants were asked to adjust their chair to an ideal position and were instructed to stay still throughout the study. The angle of eye-tracker was adjusted to make sure their eyes were captured properly.

\textbf{Measurements.} We used a 55" TV screen and subjects were seated 64" from the screen. The eye-tracker was set 20" in front of the subject. The eye-tracker was adjusted to be 12" lower than participants' eyes, and so it was angled by around 31 degrees. Figure~\ref{study_setup} is an illustration of our setup.

\textbf{Experiment.} 
To begin with, an eye-tracking calibration was performed for every participant.  
After a successful calibration, the screen displayed instructions for the experiment. 
Participants then used the space key to toggle through instruction pages that described the experiment. 

Participants were then shown a series of single face images ($1024 \times 1024$) one at a time. These images were randomly sampled from a set of real and fake images. 

Participants were asked to simply look at each image, and respond using the keyboard whether they believed an image was real or fake. As explained in the instructions, a fake image referred to an image that was generated by an AI model (the person in the image does not exist); a real image referred to an image of a person that exists in real life but it could be modified (filtered, digitally edited, etc.). Participants were supposed to press $1$ on the keyboard if they decided an image was real, and press $2$ otherwise. Halfway through the study, participants were informed of both their progress and their accuracy in distinguishing real and fake images. During the experiment, participants' eye movements, answers they selected and their reaction time for each image were recorded.

\begin{figure*}[t]
\centering
\includegraphics[width=0.80\textwidth]{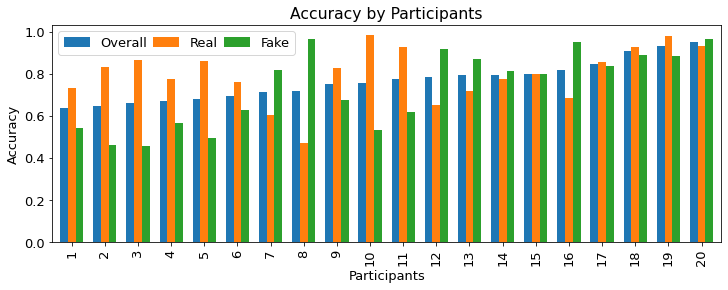
}
\caption{\textbf{Participant-wise Recognition Accuracy:} For each participant, the blue bar presents the recognition accuracy on all images; the orange bar presents the recognition accuracy on real images, and the green bar presents the recognition accuracy on fake images. Data is sorted in ascending order by overall accuracy and participant indices are reset after sorting.}
\label{acc_participants}
\end{figure*}

After a participant finished the experiment, they were asked to fill out a post-study survey regarding how they made decisions, as well as their expertise in Generative AI and image quality.

%% file: sections/3_analysis.tex
\section{Data Summary and Analysis} \label{analysis}

In this section, we present an overview of the data we collected, and show results from our data analysis.

\begin{figure}[t]
\centering
\includegraphics[width=0.4\textwidth]{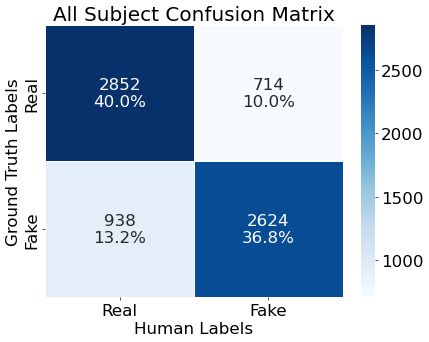}
\caption{\textbf{Confusion Matrix :} Analysis of cumulative participants performance across all real and fake images.}
\label{all_subjects_acc}
\end{figure}

\subsection{Data Summary}

In our study, we utilized real and fake face images. Real face images were randomly selected from Flickr-Faces-HQ Dataset (FFHQ)~\cite{karras2019style}, and fake face images were generated by a pre-train StyleGAN-3~\cite{karras2022stylegan3}. All the images had the same size of $1024$ x $1024$ pixels.

We recruited $20$ participants for our study. Each participant was shown $360$ images, among which $180$ were real face images and the other $180$ were fake face images. Every participant was shown a unique set of images. After filtering out invalid submissions, we had $7,128$ images with human labels and eye-tracking data for our analysis.

\subsection{Recognition Accuracy}

We analyze the per-participant accuracy in addition to the overall accuracy. 

\textbf{Overall Accuracy.} For all $20$ participants, the average recognition accuracy is $76.80$\%. To be specific, for real images, the average recognition accuracy is $79.97$\%, and for fake images, the recognition accuracy is $73.67$\%. A confusion matrix for overall participant performance is shown in Figure~\ref{all_subjects_acc}. The participants achieve a recall of $0.80$, a precision of $0.75$, and an F-1 score of $0.77$. These average scores demonstrate that in general, the participants can classify a majority of the real and fake images correctly. Besides, we do not observe a big difference between people's ability of correctly recognizing real images and correctly recognizing fake images when we consider all participants as a whole.

\begin{figure}[t]
\includegraphics[width=0.4\textwidth]{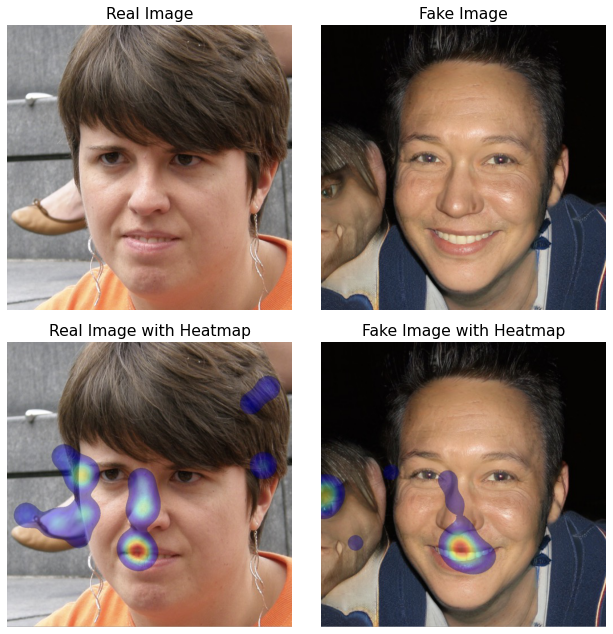}
\centering
\caption{\textbf{Gaze Heatmaps: } Real and fake face images and gaze heatmaps. Red/orange regions had longer fixation duration than blue/green regions.}
\label{heatmap}
\end{figure}

\textbf{Accuracy by Participant.} Moreover, we look at the recognition accuracy per participant.
Figure~\ref{acc_participants} shows a grouped bar plot for each participant's accuracy. Among all participants, the lowest recognition accuracy is $63.89\%$, and the highest recognition accuracy is $94.99\%$, with a median accuracy of $76.51\%$. Among all participants, six participants finished the task with an accuracy under $70\%$; nine participants finished the task with an accuracy between $70\%$ and $80\%$; two participants finished the task with an accuracy between $80\%$ and $90\%$, and three participants finished the task with an accuracy above $90\%$.

Observing the performance of each participant, we notice the following recognition patterns:

\begin{enumerate}
    \item Majority of the participants ($12$ out of $20$) have higher accuracy on correctly classifying real images than fake images.

    \item Participants who have relatively low overall accuracy (participants $1$ to $6$; overall accuracy under $70\%$) are much better at recognizing real images. \textbf{\textit{Incorrectly classifying fake images as real is the main reason why their overall accuracy is low.}}
\end{enumerate}

\subsection{Eye-tracking Data} In this subsection, we focus on analyzing eye-tracking related data. 

\begin{figure}[t]
\includegraphics[width=0.4
\textwidth]{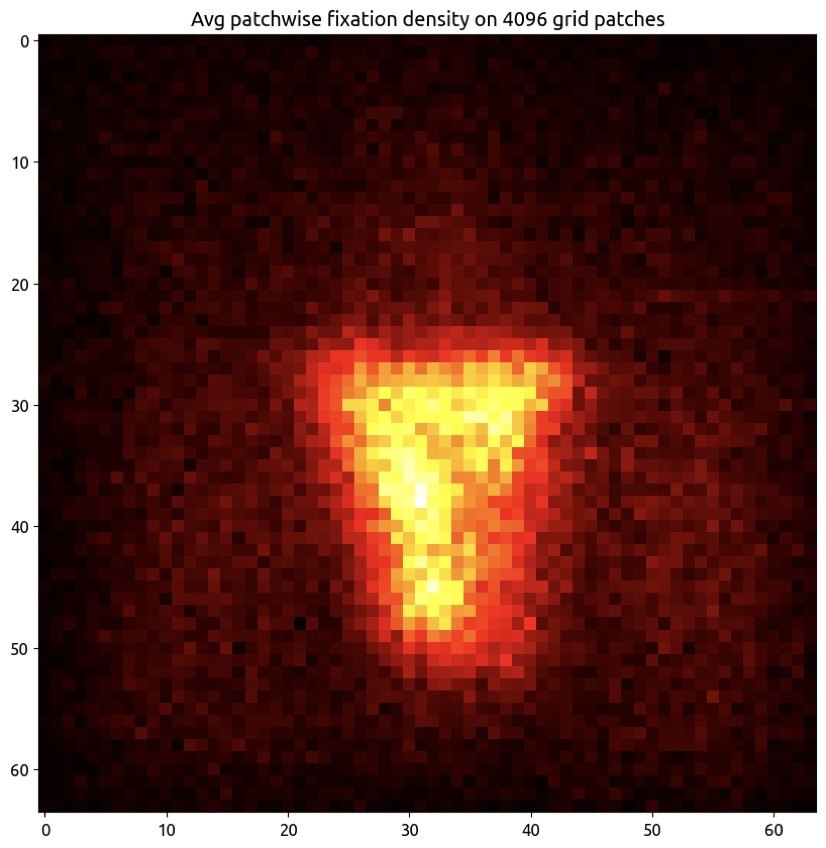}
\centering
\caption{\textbf{Plotting Human Gaze Density Spatially: } Average patch-wise fixation density on a $64 \times 64$ grid.}
\label{patch_density}
\end{figure}

\textbf{Gaze Heatmap Visualization.} To begin with, we preprocessed the raw eye-tracking data to remove invalid records and to extract all fixations and their corresponding durations. Then, we were able to plot human gaze heatmaps for better visualization. Figure~\ref{heatmap} shows examples of a real image, a fake image and their corresponding human gaze heatmaps. To plot gaze heatmaps, we utilized an open source analyzer~\footnote{https://github.com/esdalmaijer/PyGazeAnalyser}, which took an image, its corresponding fixations and duration as input. A heatmap was generated on top of the original image. The heatmap was weighted by fixation duration.

Looking at the gaze heatmaps, we find that the participants focus most of their attention on the head, facial features (eyes, nose, mouth, etc.), and any background components.

\textbf{Overall Spatial Density of Participants Gaze.} We also plot patch-wise fixation density by dividing an image into a $64 \times 64$ grid. As shown in Figure~\ref{patch_density}, participants mostly looked at the center part of an image, which is where the face is located in most of the images.

\textbf{Fixations.} In order to quantitatively understand participants' recognition patterns, we also analyze a few metrics that are related with fixations, and they include:

\begin{itemize}
    \item Number of fixations: average number of fixations in each image.
    \item Fixation duration: average duration of each fixation in seconds.
    \item Fixation convex hull area: average area of the fixation convex hull in each image. To be specific, this stands for the area of the hull that encloses all the fixation points.
    \item Fixation spatial density: dividing an image into $n$ squares, average number of fixations that fall into a square.
\end{itemize}

\begin{figure}[t]
\centering
\includegraphics[width=0.45\textwidth]{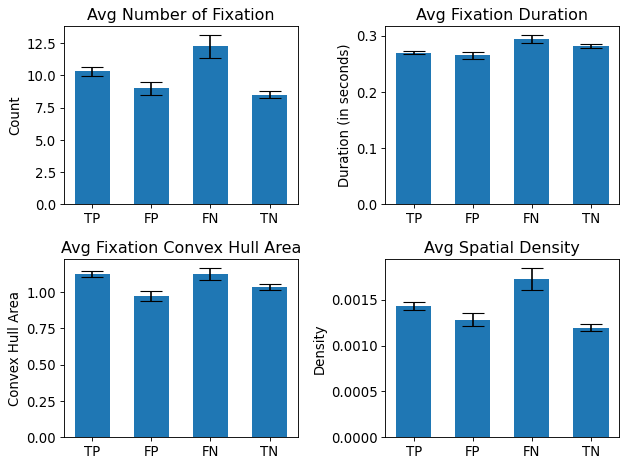}

\caption{\textbf{Analyzing Human Gaze Patterns:} Average number of fixations, average fixation duration, average convex hull area and average spatial density by TP, FP, FN and TN. Error bars stand for $95\%$ Confidence Interval.}
\label{all_fixation}
\end{figure}

Figure~\ref{all_fixation} shows plots for the 4 above mentioned metrics related with fixation by looking at 4 categories: True Positive (TP), False Positive (FP), False Negative (FN) and True Negative (TN), where real images are considered as positive and fake images are considered as negative.

Across all 4 metrics, False Negative (FN) always has the highest value, and this indicates that participants looked at some real images very carefully, but still incorrectly classified them as fake. True Negative has the smallest value in the average number of fixations and average spatial density metrics. This shows that little effort is needed from the participants to correctly classify a fake image. This may imply that some of the fake images appear to be of very low quality.

\begin{figure}[t]
\centering
\includegraphics[width=0.45\textwidth]{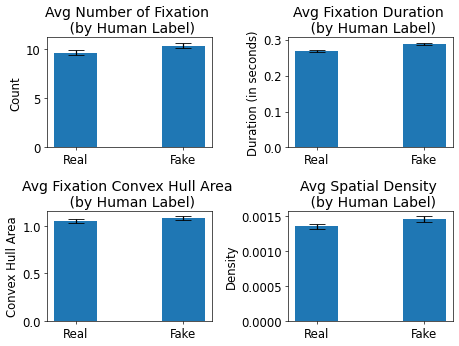}

\caption{\textbf{Analyzing Gaze Patterns based on Human Labels:  }Average number of fixations, average fixation duration, average convex hull area and average spatial density against human labels. Error bars stand for $95\%$ Confidence Interval (CI).}
\label{all_fixation_human}
\end{figure}

Figure~\ref{all_fixation_human} plots the above four metrics against the human labels, where ``Real" means an image is labeled as ``Real" by participants, and similarly, ``Fake" means an image is labeled as ``Fake" by participants. In all four subplots, the bars present the choice ``Fake" have higher average values, and it shows that participants spent more time when they suspected an image was fake.

\begin{figure}[h]
\centering
\includegraphics[width=0.45\textwidth]{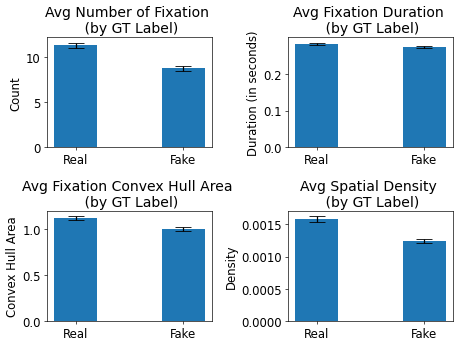}

\caption{\textbf{Analyzing Gaze Patterns based on GT Labels:  }Average number of fixations, average fixation duration, average convex hull area and average spatial density by looking at ground truth (GT) labels. Error bars stand for $95\%$ Confidence Interval (CI).}
\label{all_fixation_gt}
\end{figure}

Figure~\ref{all_fixation_gt} shows plots by looking at ground truth labels, where ``Real" means an image is an real image, and ``Fake" means an image is generated by StyleGAN-3. In all four subplots, the bars present the choice ``Real" have higher average values, and it might suggest that participants actually spent more time when an image is real.

\textbf{Figure~\ref{all_fixation_human} and Figure~\ref{all_fixation_gt} indicate that participants look at the images more carefully when they \textit{think} an image is fake. However, they actually spent more effort on real images than fake images.}




\subsection{Analysis of Different Attributes}

To better understand what features are important for human perception, we picked a few attributes that we considered can impact participants' decisions for further analysis.

\textbf{Data and labels.} We consider $5$ different attributes in this analysis, which are: whether there is a person in the image background, gender of the person in the image, skin color of the person in the image, age of the person in the image, and whether the person in the image is wearing any accessory. To be brief, they will be referred as background, gender, skin color, age and accessory below.

For this part of analysis, we used the responses from $17$ participants.
For each participants, we randomly selected $30$ real images and $30$ fake images among the $340$ images that they saw. This yielded $1020$ images in total. Then, we manually assigned them labels for the 5 attributes mentioned above and Table~\ref{tab:attributes_result} shows the semantic meaning for each numerical label in each attribute.

         

\textbf{Results.} Table~\ref{tab:attributes_result} shows the human recognition accuracy for each attribute and each label. According to these results, we find that backgrounds largely impact participants' performance on the task. Their recognition accuracy is $85\%$ when there is a person in the background, comparing to the $77\%$ when there is no person in the background. This may imply that some backgrounds were poorly generated, and participants may be able to classify such images as fake accurately. Besides, we find that participants are slightly better at recognizing males correctly comparing to females, and they are most confused by teenager images among all age level. Although we also look at skin colors and accessories in the images, we find that they both have very little impact on participants' decisions.

\begin{table*}[t]
\centering
\caption{\textbf{Analyzing Human Participant Accuracy Based on Five Attributes: } We have summarized the five attributes and the corresponding labels and their human recognition accuracy.}
\begin{tabular}{|l|c|c|c|c|c|}
\hline
\rowcolor[HTML]{D9D9D9} 
\textbf{Attributes} & \multicolumn{5}{c|}{\cellcolor[HTML]{D9D9D9}\textbf{Labels}} \\ \hline

\multirow{2}{*}{\cellcolor[HTML]{D9D9D9}\textbf{Background}} & \textbf{Present} & \textbf{Absent} & \cellcolor[HTML]{808080} & \cellcolor[HTML]{808080} & \cellcolor[HTML]{808080} \\ \cline{2-6} 
& 77\% & 85\% & \cellcolor[HTML]{808080} & \cellcolor[HTML]{808080} & \cellcolor[HTML]{808080} \\ \hline

\multirow{2}{*}{\cellcolor[HTML]{D9D9D9}\textbf{Gender}} & \textbf{Male} & \textbf{Female} & \textbf{Uncertain} & \cellcolor[HTML]{808080} & \cellcolor[HTML]{808080} \\ \cline{2-6} 
& 81\% & 77\% & 78\% & \cellcolor[HTML]{808080} & \cellcolor[HTML]{808080} \\ \hline

\multirow{2}{*}{\cellcolor[HTML]{D9D9D9}\textbf{Skin Tone}} & \textbf{Light} & \textbf{Medium} & \textbf{Dark} & \textbf{Uncertain} & \cellcolor[HTML]{808080} \\ \cline{2-6} 
& 78\% & 79\% & 79\% & 86\% & \cellcolor[HTML]{808080} \\ \hline

\multirow{2}{*}{\cellcolor[HTML]{D9D9D9}\textbf{Age}} & \textbf{Infant} & \textbf{Teen} & \textbf{Adult} & \textbf{Elderly} & \textbf{Uncertain} \\ \cline{2-6} 
& 80\% & 75\% & 80\% & 81\% & 100\% \\ \hline

\multirow{2}{*}{\cellcolor[HTML]{D9D9D9}\textbf{Accessory}} & \textbf{Present} & \textbf{Absent} & \textbf{Uncertain} & \cellcolor[HTML]{808080} & \cellcolor[HTML]{808080} \\ \cline{2-6} 
& 79\% & 78\% & 96\% & \cellcolor[HTML]{808080} & \cellcolor[HTML]{808080} \\ \hline
\end{tabular}
\label{tab:attributes_result}
\end{table*}


\begin{figure}[t]
\includegraphics[width=0.4\textwidth]{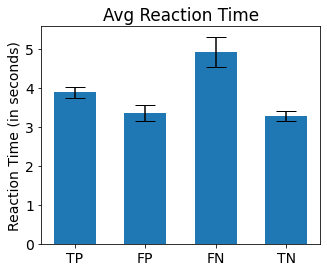}
\centering
\caption{\textbf{Analyzing Reaction Time:} Average human reaction time for 4 categories: True Positive (TP), False Positive (FP), False Negative (FN) and True Negative (TN). Error bars stand for $95\%$ Confidence Interval (CI).}
\label{avg_rt}
\end{figure}

\subsection{Reaction Time}

As mentioned in Section~\ref{intro}, we also collected human reaction time during our experiments. For each tested image, the elapsed time was recorded between when the image was first shown and when the participant responded.

Figure~\ref{avg_rt} shows the average reaction time for $4$ categories: True Positive (TP), False Positive (FP), False Negative (FN) and True Negative (TN). Among all the categories, TN has the shorted average reaction time and FN has the longest average reaction time. This is aligned with our findings in eye-tracking related metrics and indicates that some generated images are of poor quality and participants spent very little time on them. Besides, participants were confused by real images; they spent more time on some real images but decided they were fake.

\subsection{Mid-point Effect}

As mentioned in Section~\ref{design}, halfway through the study, participants were informed of both their progress and their accuracy in distinguishing real and fake images. We did this in order to study whether participants' performance changes during the experiment.

\begin{figure}[t]
\includegraphics[width=0.40\textwidth]{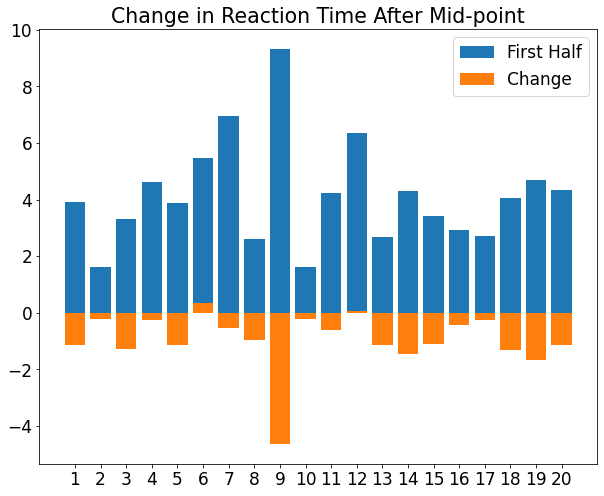}
\centering
\caption{\textbf{Analyzing Change in Reaction Time After Experiment Midpoint: }x-axis indicates participant indices, and y axis is reaction time (in seconds). Each blue bar represents a participant's average reaction time before mid-point, and the corresponding orange bar indicates the change in their reaction time. If an orange bar is above zero, that means a participant's reaction time increased for the second half. Otherwise, it means a participant's reaction time dropped for the second half.}
\label{midpoint_rt}
\end{figure}

\begin{figure}[h]
\includegraphics[width=0.40
\textwidth]{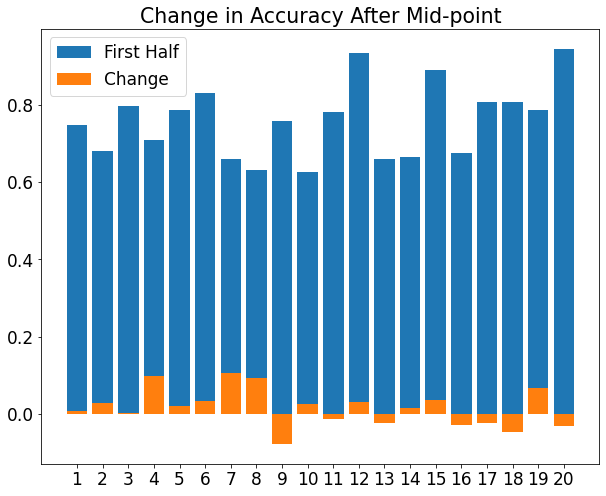}
\centering
\caption{\textbf{Analyzing Human Participant Performance Before and After Experiment Midpoint:} x-axis indicates participant indices, and y axis is recognition accuracy. Each blue bar represents a participant's accuracy before mid-point, and the corresponding orange bar indicates the change in their accuracy. If an orange bar is above zero, that means a participant's accuracy increased for the second half. Otherwise, it means a participant's accuracy dropped for the second half.}
\label{midpoint_acc}
\end{figure}

Figure~\ref{midpoint_acc} shows participants' recognition accuracy in the first half as well as the change in their accuracy after mid-point. Among all $20$ participants, $13$ participants have a higher accuracy for the second half of the task, and $7$ participants have a lower accuracy for the second half of the task. Figure~\ref{midpoint_rt} shows participants' reaction time in the first half as well as the change in their reaction time after mid-point. All $20$ participants have shorter average reaction time in the second half of the task.

We believe participants having shorter reaction times for the second half of the experiment can be attributed to fatigue. Similarly, majority participants reporting higher accuracy after the midpoint can be attributed to them becoming familiar to the task.

\subsection{Post-study Survey Results}

We also conducted a short post-study survey to gather participants' response on the following 3 questions:

\begin{enumerate}
    \item What do you look at to decide whether an image is real or fake? Options include: nose, eyes, teeth, chin, hair, ears, contour, clothes, background, others (please indicate what they are).

    \item What is your experience level in Generative AI? Rate your experience from 0 to 5; 0 means no experience at all, 5 means expert level.

    \item What is your knowledge level in image quality? Rate your experience from 0 to 5; 0 means no experience at all, 5 means expert level.
\end{enumerate}


\begin{figure}[b]
\includegraphics[width=0.40\textwidth]{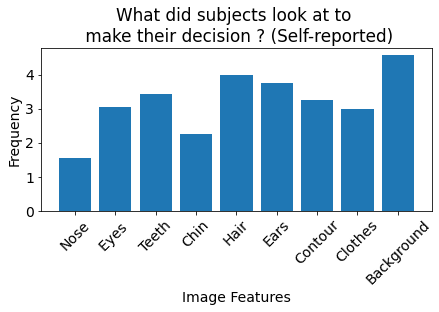}
\centering
\caption{\textbf{Self-reported Participant Responses: }Common image features participants reported they used in making their decisions; reported by participants. X-axis shows options provided in the survey, and Y-axis presents how many times these features were selected by participants.}
\label{q1}
\end{figure}

Figure~\ref{q1} shows a histogram of what participants reported as cues they used to decide whether an image was real or fake. Among all provided options, participants reported that they looked at background the most, followed by facial features including hair, ears and teeth. During the survey, some participants also provided unlisted scene elements, including: facial expression, eye orientation, accessories (eye glasses, jewelry, headgear, etc), lighting, textures on skin and clothes. This indicates that AI-generated images look real on most of the facial features, but they still have a lot to improve on the details that are not directly related with a person's face.

\begin{figure}[t]
\includegraphics[width=0.45\textwidth]{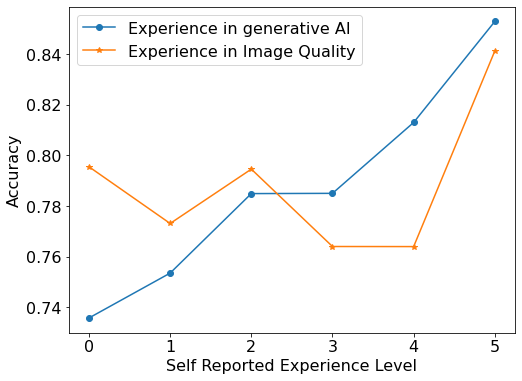}

\centering
\caption{\textbf{Self-reported Participant Responses: }Self-reported generative AI and image quality experience and corresponding accuracy scores.}
\label{q2_q3}
\end{figure}

Figure~\ref{q2_q3} shows self-reported Generative AI experience and Image Quality experience from the participants. Recognition accuracy tend to increase while experience level increased, which means having more experience in Generative AI helps participants to perform better in our task. On the other hand, it is inconclusive whether image quality background helps participants to do better.

%% file: sections/5_conclusion.tex
\section{Conclusion and Future Work} \label{conclude}

In this paper, we presented an eye-tracking psychophysical study on the human perception of real and fake face images. We collected human responses and eye-tracking data on over $7,000$ images across $20$ participants. We analyzed human recognition accuracy, as well as gaze information to explore gaze patterns that are related with human decisions. In this section, we summarize our work by presenting our findings and offering some insights on future work along this direction.

\subsection{Results Summary}

We conclude our main findings from three perspectives: human recognition accuracy, gaze patterns and other attributes.

\textbf{Recognition Accuracy.} Based on our data, the overall accuracy is $79.51\%$ for all participants. This indicates that humans are pretty good at telling what is real and what is fake even though StyleGAN-3 generates very realistic images. Besides, there is no specific difference between their accuracy on real images and fake images if we look at all participants as a whole. 

Furthermore, when we look at each participant, the lowest individual accuracy is only $63.89\%$ while the highest accuracy reached $94.99\%$. We noticed that majority of the participants ($60\%$) have higher accuracy for real images than fake images; and for those participants who have relatively low overall accuracy, incorrectly classifying fake images as real is the main reason why their accuracy is low.

\textbf{Gaze Patterns.} We visualized gaze information by plotting heatmaps on the images, and analyzed number of fixations, fixation duration, convex hall area and spatial density. 

Dividing human responses into TP, FP, FN and TN, we found that participants look at the images more carefully when
they think an image might be fake. However, they actually spent more effort on real images than fake images.

\textbf{Other Attributes.} We considered $5$ different attributes (background, gender, skin color, age and accessory) and manually labeled $1020$ images. We noticed background largely impact participants’ performance on the task, while the rest of the attributes did not impact human decisions very much.

\subsection{Discussion and Future Work}

Our study suggests that humans are pretty good at distinguishing real and fake face images, which differs from previous work \cite{lago2021more}. Analysis on specific attributes and Reaction Time showed that background is an important feature for humans to decide whether an image is real or fake, and this may indicate that StyleGAN-3 failed to generate realistic backgrounds, and may have generated images that are of bad quality and are obviously fake.

With these findings, we may consider how to improve generative models to generate better images, or how to help people to better spot fake images. For example, we could consider designing a human-in-the-loop system to improve GANs with human feedback and to develop more robust systems. Besides, investigating unique patterns in gaze pattern could involve longitudinal studies to understand how these patterns vary across different demographic groups and cultures, and furthermore to be utilized to develop customized systems based on user's preferences.